\documentclass{article}
\usepackage[accepted]{icml2005}
\usepackage{mlapa}
\usepackage{haldefs}
\usepackage{algorithm}
\usepackage{algorithmic}

\newtheorem{theorem}{Theorem}
\newtheorem{corrolary}[theorem]{Corollary}

\newcommand{\lab}[1]{$_{\textrm{#1}}$}

\newcommand{\mysection}[1]{\vspace{-3mm}\section{#1}\vspace{-1mm}}
\newcommand{\mysubsection}[1]{\vspace{-1mm}\subsection{#1}\vspace{-1mm}}
\newcommand{\mysubsubsection}[1]{\vspace{-1mm}\paragraph{#1.}}

\newenvironment{myequation}
  {\begin{small}\vspace{-2mm}\begin{equation}}%
  {\end{equation}\vspace{-4mm}\end{small}}

\allowdisplaybreaks[3]

\icmltitlerunning{Learning as Search Optimization}

\begin{document} 

\twocolumn[
\icmltitle{%
  Learning as Search Optimization: \\ %
  Approximate Large Margin Methods for Structured Prediction %
}
 
\icmlauthor{Hal Daum\'e III}{hdaume@isi.edu}
\icmlauthor{Daniel Marcu}{marcu@isi.edu}
\icmladdress{Information Sciences Institute,
             4676 Admiralty Way, Marina del Rey, CA 90292 USA}
\vskip 0.3in
]

\begin{abstract}
Mappings to structured output spaces (strings, trees, partitions,
etc.) are typically learned using extensions of classification
algorithms to simple graphical structures (eg., linear chains) in
which search and parameter estimation can be performed exactly.
Unfortunately, in many complex problems, it is rare that exact search
or parameter estimation is tractable.  Instead of learning exact
models and searching via heuristic means, we embrace this difficulty
and treat the structured output problem in terms of approximate
search.  We present a framework for learning as search optimization,
and two parameter updates with convergence theorems and bounds.
Empirical evidence shows that our integrated approach to learning and
decoding can outperform exact models at smaller computational cost.
\end{abstract} 

\vspace{-4mm}
\mysection{Introduction}

Many general techniques for learning and decoding with structured
outputs are computationally demanding, are ill-suited for dealing with
large data sets, and employ parameter optimization for an intractable
search (decoding) problem.  In some instances, such as syntactic
parsing, efficient task-specific decoding algorithms have been
developed, but, unfortunately, these are rarely applicable outside of
one specific task.

Rather than separating the learning problem from the decoding problem,
we propose to consider these two aspects in an integrated manner.  By
doing so, we are able to learn model parameters appropriate for the
search procedure, avoiding the need to heuristically combine an
\emph{a priori} unrelated learning technique and search algorithm.
After phrasing the learning problem in terms of search, we present two
online parameter update methods: a simple perceptron-style update and
an approximate large margin update.  We apply our model to two tasks:
a simple syntactic chunking task for which exact search \emph{is}
possible (to allow for comparison to exact learning and decoding
methods) and a joint tagging/chunking task for which exact search is
intractable.

\mysection{Previous Work}

Most previous work on the structured outputs problem extends standard
classifiers to linear chains.  Among these are maximum entropy Markov
models and conditional random fields
\cite{mccallum00memm,lafferty01crf}; case-factor diagrams
\cite{mcallester04cfd}; sequential Gaussian process models
\cite{altun04gp}; support vector machines for structured outputs
\cite{tsochantaridis04svmso} and max-margin Markov models
\cite{taskar03mmmn}; and kernel dependency estimation models
\cite{weston02kde}.  These models learn distributions or weights on
simple graphs (typically linear chains).  Probabilistic models are
optimized by gradient descent on the log likelihood, which requires
computable expectations of features across the structure.
Margin-based techniques are optimized by solving a quadratic program
(QP) whose constraints specify that the best structure must be
weighted higher than all other structures.  Linear chain assumptions
can reduce the exponentially-many constraints to a polynomial, but
training remains computationally expensive.

Recent effort to reduce this computational demand considers employing
constraints that the correct output only outweigh the $k$-best model
hypotheses \cite{bartlett04eg}.  Alternatively an online algorithm for
which only very small QPs are solved is also possible
\cite{mcdonald04margin}.

At the heart of all these algorithms, batch or online, likelihood- or
margin-based, is the computation:

\vspace{-4mm}
\begin{equation} \label{eq:argmax}
\hat y = \arg\max_{y \in \cY} f(x,y ; w)
\end{equation}
\vspace{-5mm}

This seemingly innocuous statement is necessary in \emph{all} models,
and ``simply'' computes the structure $\hat y$ from the set of all
possible structures $\cY$ that maximizes some function $f$ on an input
$x$, parametrized by a weight vector $w$.  This computation is
typically left unspecified, since it is ``problem specific.''

Unfortunately, this $\arg\max$ computation is, in real problems with
complex graphical structure, often intractable.  Compounding this
issue is that this best guess $\hat y$ is only one ingredient to the
learning algorithms: likelihood-based models require feature
expectations and the margin-based methods require either a $k$-best
list of best $y$ or a marginal distribution across the graphical
structure.  One alternative that alleviates some of these issues is to
use a perceptron algorithm, where \emph{only} the $\arg\max$ is
required \cite{collins02perceptron}, but performance can be adversely
affected by the fact that even the $\arg\max$ cannot be computed
exactly; see \cite{mccallum04coref} for example.

\mysection{Search Optimization}

We present the \emph{Learning as Search Optimization (LaSO)} framework
for predicting structured outputs.  The idea is to delve into
Eq~\eqref{eq:argmax} to first reduce the requirement that an algorithm
need to compute an $\arg\max$, and also to produce generic algorithms
that can be applied to problems that are significantly more complex
that the standard sequence labeling tasks that the majority of prior
work has focused on.

\mysubsection{Search}

The generic \emph{search} problem is covered in great depth in any
introductory AI book.  Its importance stems from the intractability of
computing the ``best'' solution to many problems; instead, one must
search for a ``good'' solution.  Most AI texts contain a definition of
the search problem and a general search algorithm; we work here with
that from \emcite{russell95aibook}.  A \emph{search problem} is a
structure containing four fields: \textsc{states} (the world of
exploration), \textsc{operators} (transitions in the world),
\textsc{goal test} (a subset of states) and \textsc{path cost}
(computes the cost of a path).

One defines a general search algorithm given a search problem, an
initial state and a ``queuing function.''  The search algorithm will
either fail (if it cannot find a goal state) or will return a path.
Such an algorithm (Figure~\ref{fig:search}) operates by cycling
through a queue, taking the first element off, testing it as a goal
and expanding it according to operators if otherwise.  Each node
stores the path taken to get there and the cost of this path.  The
\textit{enqueue} function places the expanded nodes, \textit{next},
onto the queue according to some variable ordering that can yield
depth-first, breadth-first, greedy, beam, hill-climbing, and A* search
(among others).  Since most search techniques can be described in this
framework, we will treat it as fixed.

\begin{figure}[t]
\small
\framebox{\hspace{-2mm}\begin{minipage}[t]{8cm}
\begin{algorithmic}
\STATE {\bf Algo} Search(\textit{problem}, \textit{initial}, \textit{enqueue})
\STATE \textit{nodes} $\leftarrow$ MakeQueue(MakeNode(\textit{problem},\textit{initial}))
\WHILE{\textit{nodes} is not empty}
\STATE \textit{node} $\leftarrow$ RemoveFront(\textit{nodes})
\STATE {\bf if} GoalTest(\textit{node}) {\bf then return} \textit{node}
\STATE \textit{next} $\leftarrow$ Operators(\textit{node})
\STATE \textit{nodes} $\leftarrow$ \textit{enqueue}(\textit{problem}, \textit{nodes}, \textit{next})
\ENDWHILE
\STATE {\bf return failure}
\end{algorithmic}
\end{minipage}\hspace{2mm}}
\caption{The generic search algorithm.}
\label{fig:search}
\vspace{-5mm}
\end{figure}

\mysubsection{Search Parameterization}

Given the search framework described, for a given task the search
problem will be fixed, the initial state will be fixed and the generic
search algorithm will be fixed.  The only place left, therefore, for
parameterization is in the \textit{enqueue} function, whose job it is
to essentially rank hypotheses on a queue.  The goal of learning,
therefore, is to produce an \textit{enqueue} function that places good
hypotheses high on the queue and bad hypotheses low on the queue.  In
the case of optimal search, this means that we will find the optimal
solution quickly; in the case of approximate search (with which we are
most interested), this is the difference between finding a good
solution or not.

In our model, we will assume that the enqueue function is based on two
components: a path component $g$ and a heuristic component $h$, and
that the score of a node will be given by $g+h$.  This formulation
includes A* search when $h$ is an admissible heuristic, heuristic
search when $h$ is inadmissible, best-first search when $h$ is
identically zero, and any variety of beam search when a queue is cut
off at a particular point at each iteration.  We will assume $h$ is
given and that $g$ is a linear function of features of the input $x$
and the path to and including the current node, $n$: $g = \vec w\T
\vec \Ph(x, n)$, where $\vec \Ph(\cdot,\cdot)$ is the vector of
features.

\mysubsection{Learning the Search Parameters}


The supervised learning problem in this search-based framework is to
take a search problem, a heuristic function, and training data with
the goal of producing a good weight vector $\vec w$ for the path
function $g$.  As in standard structured output learning, we will
assume that our training data consists of $N$-many pairs $(x^{(n)},
y^{(n)}) \in \cX \times \cY$ that tell us for a given input $x^{(n)}$
what is the correct structured output $y^{(n)}$.  We will make one
more important \emph{monotonicity assumption}: for any given node $s
\in \cS$ and an output $y \in \cY$, we can tell whether $s$ can or
cannot lead to $y$.  In the case that $s$ can lead to $y$, we refer to
$s$ as ``$y$-good.''\footnote{We assume that the \emph{loss} we
optimize is monotonic on a path in $\cS$; in this paper, we only use
$0/1$ loss.}

\begin{figure}[t]
\small
\framebox{\hspace{-2mm}\begin{minipage}[t]{8cm}
\begin{algorithmic}
\STATE {\bf Algo} Learn(\textit{problem}, \textit{initial}, \textit{enqueue}, $\vec w$, $x$, $y$)
\STATE \textit{nodes} $\leftarrow$ MakeQueue(MakeNode(\textit{problem},\textit{initial}))
\WHILE{\textit{nodes} is not empty}
\STATE \textit{node} $\leftarrow$ RemoveFront(\textit{nodes})
\IF{none of $\textit{nodes} \cup \{\textit{node}\}$ is $y$-good {\bf
    or}\\ ~~~~GoalTest(\textit{node}) and \textit{node} is not $y$-good}
\STATE \textit{sibs}  $\leftarrow$ \textit{siblings}(\textit{node}, $y$)
\STATE $\vec w \leftarrow$ \textit{update}($\vec w$, $x$, \textit{sibs}, \textit{node} $\cup$ \textit{nodes})
\STATE \textit{nodes} $\leftarrow$ MakeQueue(\textit{sibs})
\ELSE
\STATE {\bf if} GoalTest(\textit{node}) {\bf then return} $\vec w$
\STATE \textit{next} $\leftarrow$ Operators(\textit{node})
\STATE \textit{nodes} $\leftarrow$ \textit{enqueue}(\textit{problem}, \textit{nodes}, \textit{next}, $\vec w$)
\ENDIF
\ENDWHILE
\end{algorithmic}
\end{minipage}\hspace{2mm}}
\caption{The generic search/learning algorithm.}
\label{fig:learn}
\vspace{-5mm}
\end{figure}

The learning problem can thus be formulated as follows: we wish to
find a weight vector $\vec w$ such that: (1) the first goal state
dequeued is $y$-good and (2) the queue always contains at least one
$y$-good state.  In this framework, we explore an online learning
scenario, where learning is tightly entwined with the search
procedure.  From a pragmatic perspective, this makes sense: it is
useless to the model to learn parameters for cases that it will never
actually encounter.  We propose a learning algorithm of the form shown
in Figure~\ref{fig:learn}.  In this algorithm, we write
\textit{siblings}(\textit{node}, $y$) to denote the set of $y$-good
siblings of this node.  This can be calculated recursively by
back-tracing to the first $y$-good ancestor and then tracing forward
through only $y$-good nodes to the same search depth as $n$ (in tasks
where there is a unique $y$-good search path -- which is common -- the
sibling of a node is simply the appropriate initial segment of this
path).

There are two changes to the search algorithm to facilitate learning
(comparing Figure~\ref{fig:search} and Figure~\ref{fig:learn}).  The
first change is that whenever we make an error (a non $y$-good goal
node is dequeued or none of the queue is $y$-good), we update the
weight vector $\vec w$.  Secondly, when an error is made, instead of
continuing along this bad search path, we instead clear the queue and
insert all the \emph{correct} moves we could have
made.\footnote{Performing parameter optimization within search
resembles reinforcement learning without the confounding factor of
``exploration.''  Early research in reinforcement learning focused on
arbitrary input/output mappings \cite{farley54selforganizing}, though
this was not framed as search.  Later, \emph{associative RL} was
introduced, where a \emph{context input} (akin to our input $x$) was
given to a RL algorithm \cite{barto81context,barto85rl}.  Similar
approaches attempt to predict value functions for generalization using
techniques such as temporal difference (TD) or Q-learning
\cite{bellman63dp,boyan96evaluation,sutton96generalization}.  More
recently, \emcite{zhang97combinatorial} applied RL techniques to
solving combinatorial scheduling problems, but again focus on the
standard TD($\la$) framework.  These frameworks, however, are not
explicitly tailored for supervised learning and without the aid of our
monotonicity assumption it is difficult to establish convergence and
generalization proofs.  Despite these differences, our search
optimization framework clearly lies on the border between supervised
learning and reinforcement learning, and further investigation may
reveal interesting connections.  }

Note that this algorithm cannot fail (in the sense that it will always
find a goal state).  Aiming at a contradiction, suppose it were to
fail; this would mean that \textit{nodes} would have become empty.
Since ``Operators'' will never return an empty set, this means that
\textit{sibs} must have been empty.  But since a node that is inserted
into the queue is either itself good or has an ancestor that is good,
so could never have become empty.  (There may be a complication with
cyclic search spaces -- in this case, both algorithms need to be
augmented with some memory to avoid such loops, as is standard.)

\mysubsection{Parameter Updates}

We propose two methods for updating the model parameters.  To
facilitate discussion, we will refer to a problem as \emph{linearly
separable} if there exists a weight vector $\vec w$ with $\norm{\vec
w}_2 \leq 1$ such that the search algorithm parameterized by $\vec w$
(a) will not fail and (b) will return an optimal solution.  Note that
with this definition, linear separability is a joint property of the
problem \emph{and} the search algorithm: what is separable with exact
search may not be separable with a heuristic search.  In the case of
linearly separable data, we define the \emph{margin} as the maximal
$\ga$ such that the data remain separable when all $y$-good states are
down-weighted by $\ga$.  In other words, $\ga$ is the minimum over all
decisions of $\max_{g,b} |\vec w\T \vec \Ph(x,g) - \vec w\T \vec
\Ph(x,b)|$, where $g$ is a $y$-good node and $b$ is a $y$-bad node.

\mysubsubsection{Perceptron Updates}

A simple perceptron-style update rule \cite{rosenblatt58perceptron},
given $(\vec w, x, \textit{sibs}, \textit{nodes})$ is $\vec w
\leftarrow \vec w + \vec \De$, where:


\begin{myequation} \label{eq:delta}
\vec \De = 
       \sum_{n \in \textit{sibs}}  \frac {\vec \Ph(x, n)} {|\textit{sibs}|} -
       \sum_{n \in \textit{nodes}} \frac {\vec \Ph(x, n)} {|\textit{nodes}|}
\end{myequation}

When an update is made, the feature vector for the incorrect decisions
are subtracted off, and the feature vectors for all possible correct
decisions are added.  Whenever $|\textit{sibs}|=|\textit{nodes}|=1$,
this looks exactly like the standard perceptron update.  When there is
only one sibling but many nodes, this resembles the gradient of the
log likelihood for conditional models after approximating the ``log
sum exp'' with Jensen's inequality to turn it into a simple sum.  When
there is more than one correct next hypothesis, this update rule
resembles that used in multi-label or ranking variants of the
perceptron \cite{crammer03rank}.  In that work, different
``weighting'' schemes are proposed, including, for instance, one that
weights the nodes in the sums proportional to the loss suffered; such
schemes are also possible in our framework, but space prohibits a
discussion of them here.  Based on this update, we can prove the
following theorem:

\begin{theorem}
For any training sequence that is separable with by a margin of size
$\ga$, using the perceptron update rule the number of errors made
during training is bounded above by $R^2/\ga^2$, where, $R$ is a
constant such that for all training instances $(x,y)$, for all nodes
$n$ in the path to $y$ and all successors $m$ of $n$ (good or
otherwise), $\norm{\vec \Ph(x,n) - \vec \Ph(x,m)}_2 \leq R$.
\end{theorem}

In the case of inseparable data, we follow \emcite{freund99perceptron}
and define $D_{\vec w,\ga}$ as the least obtainable error with weights
$\vec w$ and margin $\ga$ over the training data: $D_{\vec w,\ga} =
[\sum_s (\max \{0, \ga - p_s\})^2]^{1/2}$, where the sum is over all
states leading to a solution and $p_s$ is the empirical margin between
the correct state and the hypothesized state $s$.  Using this
notation, we obtain two corollaries (proofs are direct adaptations of
\emcite{freund99perceptron} and \emcite{collins02perceptron}):

\begin{corrolary}
For \emph{any} training sequence, the number of mistakes made by the
training algorithm is bounded above by $\min_{\vec w,\ga} (R+D_{\vec
w,\ga})^2 / \ga^2$, where $R$ is as before.
\end{corrolary}

\begin{corrolary}  \label{cor:perc-gen}
For any i.i.d. training sequence of length $n$ and any test example
$(\tilde x, \tilde y)$, the probability of error on the test example
is bounded above by $(2/(n+1)) \Ep\{\min_{\vec w,\ga} (R+D_{\vec
  w,\ga})^2 / \ga^2\}$, where the expectation is taken over all $n+1$
data points.
\end{corrolary}

\mysubsubsection{Approximate Large Margin Updates}

One major disadvantage to the perceptron algorithm is that it only
updates the weights when errors are made.  This can lead to a brittle
estimate of the parameters, in which the ``good'' states are weighted
only minimally better than the ``bad'' states.  We would like to
enforce a large margin between the good states and bad states, but
would like to do so without adding significant computational
complexity to the problem.  In the case of binary classification,
\emcite{gentile-alma} has presented an online, approximate large
margin algorithm that trains similarly to a perceptron called ALMA.
The primary difference (aside from a step size on the weight updates)
is that updates are made if either (a) the algorithm makes a mistake
or (b) \emph{the algorithm is close to making a mistake.}  Here, we
adapt this algorithm to structured outputs in our framework.



Our algorithm, like ALMA, has four parameters: $\al,B,C,p$.  $\al$
determines the degree of approximation required: for $\al=1$, the
algorithm seeks the true maximal margin solution, for $\al<1$, it
seeks one within $\al$ of the maximal.  $B$ and $C$ can be seen as
tuning parameters, but a default setting of $B=1/\al$ and $C=\sqrt 2$
is reasonable (see Theorem 4 below).  We measure the instance vectors
with norm $p$ and the weight vector with its dual value $q$ (where
$1/p+1/q=1$).  We use $p=q=2$, but large $p$ produces sparser
solutions, since the weight norm will approach $1$.  The update is:

\begin{myequation}
\vec w \leftarrow \wp(\vec w + Ck^{-1/2}~\wp(\De))
\end{myequation}

\noindent Here, $k$ is the ``generation'' number of the weight vector
(initially 1 and incremented at every update) and $\wp(\vec u)$ is the
projection of $\vec u$ into the $l_2$ unit sphere: $\vec u / \max \{
1, \norm{\vec u}_2 \}$.  One final change to the algorithm is to
down-weight the score of all $y$-good nodes by $(1-\al)Bk^{-1/2}$.
Thus, a good node will only survive if it is good by a large margin.
This setup gives rise to a bound on the number of updates made (proof
sketched in Appendix A) and two corollaries (proofs are nearly
identical to Theorem 4 and \cite{gentile-alma}):

\begin{theorem} \label{thm:alma}
For any training sequence that is separable with by a margin of size
$\ga$ using the approximate large margin update rule with parameters
$\al, B=\sqrt 8 / \al, C=\sqrt 2$, the number of errors made during
training is bounded above by $\frac 2 {\ga^2} \left(\frac 2 \al -
1\right)^2 + \frac 8 \al - 4$.
\end{theorem}

\begin{corrolary}
Suppose for a given $\al$, $B$ and $C$ are such that $C^2 + 2(1-\al)BC
= 1$; letting $\rh = (C\ga)^{-2}$, the number of corrections made is
bounded above by:

\begin{myequation} \label{eq:nonlin-bounds}
\min_{\vec w,\ga}
\frac 1 \ga D_{\vec w,\ga} + \frac {\rh^2} 2 + \rh
  \left[
    \frac {\rh^2} 4 + \frac 1 \ga D_{\vec w,\ga} + 1
  \right]^{1/2}
\end{myequation}
\end{corrolary}

\begin{corrolary} \label{cor:alma-gen}
For any i.i.d. training sequence of length $n$ and any test example
$(\tilde x, \tilde y)$, the probability of error on the test example
is bounded above by $(2/(n+1)) \Ep\{\cdot\}$, where $(\cdot)$ is given
in Eq~\eqref{eq:nonlin-bounds} and the expectation is taken over all
$n+1$ data points.
\end{corrolary}

\mysection{Experiments}
\vspace{3mm}

\mysubsection{Syntactic Chunking} \label{sec:chunking}

The syntactic chunking problem is a sequence segmentation and labeling
problem; for example:

\begin{quote}\vspace{-2mm}
\tiny
[Great American]\lab{NP}
[said]\lab{VP}
[it]\lab{NP}
[increased]\lab{VP}
[its loan-loss reserves]\lab{NP}
[by]\lab{PP}
[\$ 93 million]\lab{NP}
[after]\lab{PP}
[reviewing]\lab{VP}
[its loan portfolio]\lab{NP}
,
[raising]\lab{VP}
[its total loan and real estate reserves]\lab{NP}
[to]\lab{PP}
[\$ 217 million]\lab{NP}
.
\vspace{-4mm}\end{quote}

Typical approaches to this problem recast it as a sequence labeling
task and then solve it using any of the standard sequence labeling
models; see \cite{sha02crf} for a prototypical example using CRFs.
The reduction to sequence labeling is typically done through the
``BIO'' encoding, where the beginning of an $X$ phrase is tagged
B-$X$, the non-beginning (inside) of an $X$ phrase is tagged I-$X$ and
any word not in a phrase is tagged O (outside).  More recently,
\emcite{sarawagi04scrf} have described a straightforward extension to
the CRF (called a Semi-CRF) in which the segmentation and labeling is
done directly.

We explore similar models in the context of syntactic chunking, where
entire chunks are hypothesized, and no reduction to word-based labels
is made.  We use the same set of features across all models, separated
into ``base features'' and ``meta features.''  The base features apply
to words individually, while meta features apply to entire chunks.
The base features we use are: the chunk length, the word (original,
lower cased, stemmed, and original-stem), the case pattern of the
word, the first and last 1, 2 and 3 characters, and the part of speech
and its first character.  We additionally consider membership features
for lists of names, locations, abbreviations, stop words, etc.  The
meta features we use are, for any base feature $b$, $b$ at position
$i$ (for any sub-position of the chunk), $b$ before/after the chunk,
the entire $b$-sequence in the chunk, and any 2- or 3-gram tuple of
$b$s in the chunk.  We use a first order Markov assumption (chunk
label only depends on the most recent previous label) and all features
are placed on labels, not on transitions.  In this task, the
$\arg\max$ computation from Eq~\eqref{eq:argmax} \emph{is} tractable;
moreover, through a minor adaptation of the standard HMM forward and
backward algorithms, we can compute feature expectations, which enable
us to do training in a likelihood-based fashion.

Our search space is structured so that each state is the segmentation
and labeling of an initial segment of the input string, and an
operation extends a state by an entire labeled chunk (of any number of
words).  For instance, on the example shown at the beginning of this
section, the initial hypothesis would be empty; the first correct
child would be to hypothesize a chunk of length $2$ with the tag NP.
The next correct hypothesis would be a chunk of length $1$ with tag
VP.  This process would continue until the end of the sentence is
reached.  For beam search, we execute search as described, but after
every expansion we only retain the $b$ best hypotheses to continue on
to the next round.

Our models for this problem are denoted \textsc{LaSOp}$_b$ and
\textsc{LaSOa}$_b$, where $b$ is the size of the beam we use in
search, which we vary over $\{ 1, 5, 25, \infty \}$, where $\infty$
denotes full, exact Viterbi search and forward-backward updates
similar to those used in the semi-CRF.  This points out an important
issue in our framework: if the graphical structure of the problem
\emph{is} amenable to exact search and exact updates, then the
framework can accommodate this.  In this case, for example, when using
exact search, updates are only made at the end of decoding when the
highest ranking output is incorrect (after adjusting the weights down
for \textsc{LaSOa}), but other than this exception, the sum over the
bad nodes in the updates is computed over the entire search lattice
and strongly resemble almost identical to those used in the
conditional likelihood models for the gradient of the log
normalization constant.  We always use averaged weights.

We report results on the CoNLL 2000 data set, which includes $8936$
training sentences ($212k$ words) and $2012$ test sentences ($47k$
words).  We compare our proposed models against several baselines.
The first baseline is denoted \textsc{ZDJ02} and is the best system on
this task to date \cite{zhang02chunking}.  The second baseline is the
likelihood-trained model, denoted \textsc{SemiCRF}.  We use $10\%$ of
the training data to tune model parameters.  The third baseline is the
standard structured perceptron algorithm, denoted \textsc{Perceptron}.
For the \textsc{SemiCRF}, this is the prior variance; for the online
algorithms, this is the number of iterations to run (for ALMA, $\al =
0.9$; changing $\al$ in the range $[0.5,1]$ does not affect the score
by more than $\pm 0.1$ in all cases).

The results, in terms of training time, test decoding time, precision,
recall and f-score are shown in Table~\ref{tab:chunking}.  As we can
see, the \textsc{SemiCRF} is by far the most computationally expensive
algorithm, more than twice as slow to train than even the
\textsc{LaSOp}$_{\infty}$ algorithm.  The \textsc{Perceptron} has
roughly comparable training time to the exactly trained \textsc{LaSO}
algorithms (slightly faster since it only updates for the best
solution), but its performance falls short.  Moreover, decoding time
for the \textsc{SemiCRF} takes a half hour for the two thousand test
sentences, while the greedy decoding takes only 52 seconds.  It is
interesting to note that at the larger beam sizes, the large margin
algorithm is actually faster than the perceptron algorithm.

\begin{table}[t]
\small
\center
\caption{Results on syntactic chunking task; columns are training and
  testing time (h:m), and precision/recall/f-score on test data.}
\label{tab:chunking}
\vspace{3mm}
\begin{tabular}{|@{~}l@{~}||@{~}r@{~}|@{~}r@{~}||@{~}c@{~}|@{~}c@{~}|@{~}c@{~}|}
\hline
 & {\bf Train} & {\bf Test} & {\bf Pre} & {\bf Rec} & {\bf F} \\
\hline
\textsc{ZDJ02}            & -     & -     & 94.3 & 94.0 & 94.1 \\
\hline
\textsc{SemiCRF}          & 53:56 &   :31 & 92.3 & 92.1 & 92.2 \\
\hline
\textsc{Perceptron}       & 18:05 &   :22 & 93.4 & 93.5 & 93.4 \\
\hline
\textsc{LaSOp}$_{1}$      &   :55 &   :01 & 92.5 & 92.3 & 92.4 \\
\textsc{LaSOp}$_{5}$      &  1:49 &   :04 & 93.7 & 92.6 & 93.1 \\
\textsc{LaSOp}$_{25}$     &  6:32 &   :11 & 94.2 & 94.1 & 94.1 \\
\textsc{LaSOp}$_{\infty}$ & 21:43 &   :24 & 94.3 & 94.1 & 94.2 \\
\hline
\textsc{LaSOa}$_{1}$      &   :51 &   :01 & 93.6 & 92.5 & 93.0 \\
\textsc{LaSOa}$_{5}$      &  2:04 &   :04 & 93.8 & 94.4 & 94.3 \\
\textsc{LaSOa}$_{25}$     &  5:59 &   :10 & 93.9 & 94.6 & 94.4 \\
\textsc{LaSOa}$_{\infty}$ & 20:12 &   :25 & 94.0 & 94.8 & 94.4 \\
\hline
\end{tabular}
\vspace{-6mm}
\end{table}

In terms of the quality of the output, the \textsc{SemiCRF} falls
short of the previous reported results (92.2 versus 94.1 f-score).
Our simplest model, \textsc{LaSOp}$_1$ already outperforms the
\textsc{SemiCRF} with an f-score of 92.4; the large margin variant
achieves 93.0.  Increasing the beam past $5$ does not seem to help
with large margin updates, where performance only increases from
$94.3$ to $94.4$ going from a beam of 5 to an infinite beam (at the
cost of an extra 18 hours of training time).


\mysubsection{Joint Tagging and Chunking} \label{sec:joint}

In Section~\ref{sec:chunking}, we described an approach to chunking
based on search without reduction.  This assumed that part of speech
tagging had been performed as a pre-processing step.  In this section,
we discuss models in which part of speech tagging and chunking are
performed jointly.  This task has previously been used as a benchmark
for factorized CRFs \cite{sutton04fcrfs}.  In that work, the authors
discuss many approximate inference methods to deal with the fact that
inference in such joint models is intractable.

For this task, we \emph{do} use the BIO encoding of the chunks so that
a more direct comparison to the factorized CRFs would be possible.  We
use the same features as the last section, together with the regular
expressions given by \cite{sutton04fcrfs} (so that our feature set and
their feature set are nearly identical).  We do, however, omit their
final feature, which is active whenever the part of speech at position
$i$ matches the most common part of speech assigned by Brill's tagger
to the word at position $i$ in a very large corpus of tagged data.
This feature is somewhat unrealistic: the CoNLL data set is a small
subset of the Penn Treebank, but the Brill tagger is trained on all of
the Treebank.  By using this feature, we are, in effect, able to
leverage the rest of the Treebank for part of speech tagging.  Using
just their features without the Brill feature, our performance is
quite poor, so we added the lists described in the previous section.

In this problem, states in our search space are again initial taggings
of sentences (both part of speech tags and chunk tags), but the
operators simply hypothesize the part of speech and chunk tag for the
single next word, with the obvious constraint that an I-$X$ tag cannot
follow anything but a B-$X$ or I-$X$ tag.

\begin{table}[t]
\small
\center
\caption{Results on joint tagging/chunking task; columns are time to
  train (h:m), tag accuracy, chunk accuracy, joint accuracy and chunk
  f-score.}
\label{tab:joint}
\vspace{3mm}
\begin{tabular}{|@{~}l@{~}||@{~}r@{~}|@{~}r@{~}||@{~}c@{~}|@{~}c@{~}|@{~}c@{~}||@{~}c@{~}|}
\hline
 & {\bf Train} & {\bf Test} & {\bf Tag} & {\bf Chn} & {\bf Jnt} & {\bf F}\\
\hline
\textsc{Sutton}           & -     & -     & 98.9 & 97.4 & 96.5 & 93.9 \\
\hline
\textsc{LaSOp}$_{1}$      &  1:29 &   :01 & 98.9 & 95.5 & 94.7 & 93.1 \\
\textsc{LaSOp}$_{5}$      &  3:24 &   :04 & 98.9 & 95.8 & 95.1 & 93.5 \\
\textsc{LaSOp}$_{10}$     &  4:47 &   :09 & 98.9 & 95.9 & 95.1 & 93.5 \\
\textsc{LaSOp}$_{25}$     &  4:59 &   :16 & 98.9 & 95.9 & 95.1 & 93.7 \\
\textsc{LaSOp}$_{50}$     &  5:53 &   :30 & 98.9 & 95.8 & 94.9 & 93.4 \\
\hline
\textsc{LaSOa}$_{1}$      &   :41 &   :01 & 99.0 & 96.5 & 95.8 & 93.9 \\
\textsc{LaSOa}$_{5}$      &  1:43 &   :03 & 99.0 & 96.8 & 96.1 & 94.2 \\
\textsc{LaSOa}$_{10}$     &  2:21 &   :07 & 99.1 & 97.3 & 96.4 & 94.4 \\
\textsc{LaSOa}$_{25}$     &  3:38 &   :20 & 99.1 & 97.4 & 96.6 & 94.3 \\
\textsc{LaSOa}$_{50}$     &  3:15 &   :23 & 99.1 & 97.4 & 96.6 & 94.4 \\
\hline
\end{tabular}
\vspace{-7mm}
\end{table}

The results are shown in Table~\ref{tab:joint}.  The models are
compared against \textsc{Sutton}, the factorized CRF with tree
reparameterization.  We do not report on infinite beams, since such a
calculation is intractable.  We report training time\footnote{Sutton
et al. report a training time of 13.6 hours on 5\% of the data (400
sentences); it is unclear from their description how this scales.  The
scores reported from their model are, however, based on training on
the full data set.}, testing time, tag accuracy, chunk accuracy and
joint accuracy and f-score for chunking.  In this table, we can see
that the large margin algorithms are much faster to train than the
perceptron (they require fewer iterations to converge -- typically two
or three compared to seven or eight).  In terms of chunking f-score,
none of the perceptron-style algorithms is able to out-perform the
\textsc{Sutton} model, but our \textsc{LaSOa} algorithms easily
outperform it.  With a beam of only 1, we achieve the same f-score
(93.9) and with a beam of 10 we get an f-score of 94.4.  Comparing
Table~\ref{tab:chunking} and Table~\ref{tab:joint}, we see that, in
general, we can do a better job chunking with the large margin
algorithm when we do part of speech tagging simultaneously.

To verity Theorem 4 experimentally, we have run the same experiments
using a $1000$ sentence ($25k$ word) subset of the training data (so
that a positive margin could be found) with a beam of $5$.  On this
data, \textsc{LaSOa} made $15932$ corrections.  The empirical margin
at convergence was $1.299e-2$; according to Theorem 4, the number of
updates should have been $\leq 17724$, which is borne out
experimentally.

\mysubsection{Effect of Beam Size}

Clearly, from the results presented in the preceding sections, the
beam size plays an important role in the modeling.  In many problems,
particularly with generative models, training is done exactly, but
decoding is done using an inexact search.  In this paper, we have
suggested that learning and decoding should be done in the \emph{same}
search framework, and in this section we briefly support this
suggestion with empirical evidence.  For our experiments, we use the
joint tagging/chunking model from Section~\ref{sec:joint} and
experiment by independently varying the beam size for \emph{training}
and the beam size for \emph{decoding}.  We show these results in
Table~\ref{tab:beam}, where the training beam size runs vertically and
the decoding beam size runs horizontally; the numbers we report are
the chunk f-score.

In these results, we can see that the diagonal (same training beam
size as testing beam size) is heavy, indicating that training and
testing with the same beam size is useful.  This difference is
particularly strong when one of the sizes is $1$ (i.e., pure greedy
search is used).  When training with a beam of one, decoding with a
beam of 5 drops f-score from $93.9$ (which is respectable) to $90.5$
(which is poor).  Similarly, when a beam of one is used for decoding,
training with a beam of 5 drops performance from $93.9$ to $92.8$.
The differences are less pronounced with beams $\geq 10$, but the
trend is still evident.  We believe (without proof) that when the beam
size is large enough that the loss incurred due to search errors is at
most the loss incurred due to modeling errors, then using a different
beam for training and testing is acceptable.  However, when some
amount of the loss is due to search errors, then a large part of the
learning procedure is aimed at learning how to \emph{avoid} search
errors, not necessarily modeling the data.  It is in these cases that
it is important that the beam sizes match.

\begin{table}[t]
\small
\center
\caption{Effect of beam size on performance; columns are for constant
  decoding beam; rows are for constant training beam.  Numbers are
  chunk f-score on the joint task.}
\label{tab:beam}
\vspace{3mm}
\begin{tabular}{|r|c|c|c|c|c|}
\hline
         & {\bf  1} & {\bf  5} & {\bf 10} & {\bf 25} & {\bf 50} \\ \hline
{\bf  1} &  \em 93.9    &      92.8    &      91.9    &      91.3    &      90.9    \\ \hline
{\bf  5} &      90.5    &  \em 94.3    &  \em 94.4    &      94.1    &      94.1    \\ \hline
{\bf 10} &      89.6    &  \em 94.3    &  \em 94.4    &      94.2    &      94.2    \\ \hline
{\bf 25} &      88.7    &      94.2    &  \em 94.5    &  \em 94.3    &      94.3    \\ \hline
{\bf 50} &      88.4    &      94.2    &  \em 94.4    &      94.2    &  \em 94.4    \\ \hline
\end{tabular}
\vspace{-5mm}
\end{table}

\mysection{Summary and Discussion}

In this paper, we have suggested that one views the learning with
structured outputs problem as a search optimization problem and that
the same search technique should be applied during both learning and
decoding.  We have presented two parameter update schemes in the LaSO
framework, one perceptron-style and the other based on an approximate
large margin scheme, both of which can be modified to work in kernel
space or with alternative norms (but not both).

Our framework most closely resembles that used by the incremental
parser of \emcite{collins04incremental}.  There are, however, several
differences between the two methodologies.  Their model builds on
standard perceptron-style updates \cite{collins02perceptron} in which
a full pass of decoding is done before any updates are made, and thus
does not fit into the search optimization framework we have outlined.
Collins and Roark found experimentally that stopping the parsing early
whenever the correct solution falls out of the beam results in
drastically improved performance.  However, theyhad little theoretical
justification for doing so.  These ``early updates,'' however, do
strongly resemble our update strategy, with the difference that when
Collins and Roark make an error, they stop decoding the current input
and move on to the next; on the other hand, when our model makes an
error, it continues from the correct solution(s).  This choice is
justified both theoretically and experimentally.  On the tasks
reported in this paper, we observe the same phenomenon: early updates
are better than no early updates, and the search optimization
framework is better than early updates.  For instance, in the joint
tagging/chunking task from Section~\ref{sec:joint}, using a beam of
10, we achieved an f-score of $94.4$ in our framework; using only
early updates, this drops to $93.1$ and using standard perceptron
updates, it drops to $92.5$.

Our work also bears a resemblance to training \emph{local classifiers}
and combining them together with global inference
\cite{punyakanok01inference}.  The primary difference is that when
learning local classifiers, one must assume to have access to all
possible decisions and must rank them according to some loss function.
Alternatively, in our model, one only needs to consider alternatives
that are in the queue at any given time, which gives us direct access
to those aspects of the search problem that are easily confused.
This, in turn, resembles the online large margin algorithms proposed
by \emcite{mcdonald04margin}, which suffer from the problem that the
$\arg\max$ must be computed exactly.  Finally, one can also consider
our framework in the context of game theory, where it resembles the
iterated gradient ascent technique described by \emcite{kearns00iga}
and the closely related marginal best response framework
\cite{zinkevich05marginal}.

We believe that LaSO provides a powerful framework to learn to predict
structured outputs.  It enables one to build highly effective models
of complex tasks efficiently, without worrying about how to normalize
a probability distribution, compute expectations, or estimate
marginals.  It necessarily suffers against probabilistic models in
that the output of the classifier will not be a probability; however,
in problems with exponential search spaces, normalizing a distribution
is quite impractical.  In this sense, it compares favorably with the
energy-based models proposed by, for example, \emcite{lecun05loss},
which also avoid probabilistic normalization, but still require the
exact computation of the $\arg\max$.  We have applied the model to two
comparatively trivial tasks: chunking and joint tagging/chunking.
Since LaSO is not limited to problems with clean graphical structures,
we believe that this framework will be appropriate for many other
complex structured learning problems.

\begin{footnotesize}

\vspace{-1mm}
\subsection*{Acknowledgments}

We thank Michael Collins, Andrew McCallum, Charles Sutton, Thomas
Dietterich, Fernando Pereira and Ryan McDonald for enlightening
discussions, as well as the anonymous reviewers who gave very helpful
comments.  This work was supported by DARPA-ITO grant
NN66001-00-1-9814 and NSF grant IIS-0326276.

\nocite{}

\vspace{-3mm}

\begin{thebibliography}{}

\vspace{-1.2mm}\bibitem[Altun et~al.\/, 2004][Altun et~al.\/][2004]{altun04gp}
Altun, Y., Hofmann, T., \& Smola, A. (2004).
\newblock Gaussian process classification for segmenting and annotating
  sequences.
\newblock {\em ICML}.

\vspace{-2mm}\bibitem[Bartlett et~al.\/, 2004][Bartlett et~al.\/][2004]{bartlett04eg}
Bartlett, P.~L., Collins, M., Taskar, B., \& McAllester, D. (2004).
\newblock Exponentiated gradient algorithms for large-margin structured
  classification.
\newblock {\em NIPS}.

\vspace{-2mm}\bibitem[Barto \& Anandan, 1985][Barto and Anandan][1985]{barto85rl}
Barto, A., \& Anandan, P. (1985).
\newblock Pattern recognizing stochastic learning automata.
\newblock {\em IEEE Trans. Systems, Man and Cybernetics}.

\vspace{-2mm}\bibitem[Barto et~al.\/, 1981][Barto et~al.\/][1981]{barto81context}
Barto, A., Sutton, R., \& Watkins, C. (1981).
\newblock Associative search network: A reinforcement learning associative
  memory.
\newblock {\em Biological Cybernetics}.

\vspace{-2mm}\bibitem[Bellman et~al.\/, 1963][Bellman et~al.\/][1963]{bellman63dp}
Bellman, R., Kalaba, R., \& Kotkin, B. (1963).
\newblock Polynomial approximation -- a new computational technique in dynamic
  programming: Allocation processes.
\newblock {\em Mathematics of Computation}

\vspace{-2mm}\bibitem[Boyan \& Moore, 1996][Boyan and Moore][1996]{boyan96evaluation}
Boyan, J., \& Moore, A.~W. (1996).
\newblock Learning evaluation functions for large acyclic domains.
\newblock {\em ICML}.

\vspace{-2mm}\bibitem[Collins, 2002][Collins][2002]{collins02perceptron}
Collins, M. (2002).
\newblock Discriminative training methods for hidden {Markov} models: Theory
  and experiments with perceptron algorithms.
\newblock {\em EMNLP}.

\vspace{-2mm}\bibitem[Collins \& Roark, 2004][Collins and Roark][2004]{collins04incremental}
Collins, M., \& Roark, B. (2004).
\newblock Incremental parsing with the perceptron algorithm.
\newblock {\em ACL}.

\vspace{-2mm}\bibitem[Crammer \& Singer, 2003][Crammer and Singer][2003]{crammer03rank}
Crammer, K., \& Singer, Y. (2003).
\newblock A family of additive online algorithms for category ranking.
\newblock {\em JMLR}, {\em 3}.

\vspace{-2mm}\bibitem[Farley \& Clark, 1954][Farley and Clark][1954]{farley54selforganizing}
Farley, B., \& Clark, W. (1954).
\newblock Simulation of self-organizing systems by digital computer.
\newblock {\em I.R.E Trans. Information Theory}.

\vspace{-2mm}\bibitem[Freund \& Shapire, 1999][Freund and Shapire][1999]{freund99perceptron}
Freund, Y., \& Shapire, R. (1999).
\newblock Large margin classification using the perceptron algorithm.
\newblock {\em ML}, {\em 37}.

\vspace{-2mm}\bibitem[Gentile, 2001][Gentile][2001]{gentile-alma}
Gentile, C. (2001).
\newblock A new approximate maximal margin classification algorithm.
\newblock {\em JMLR}, {\em 2}.

\vspace{-2mm}\bibitem[Lafferty et~al.\/, 2001][Lafferty et~al.\/][2001]{lafferty01crf}
Lafferty, J., McCallum, A., \& Pereira, F. (2001).
\newblock Conditional random fields: Probabilistic models for segmenting and
  labeling sequence data.
\newblock {\em ICML}.

\vspace{-2mm}\bibitem[Kearns et~al.\/, 2000][Kearns et~al.\/][2000]{kearns00iga}
Kearns, M., Mansour, Y., \& Singh, S. (2000).
\newblock Nash convergence of gradient dynamics in General-Sum games.
\newblock {\em UAI}.

\vspace{-2mm}\bibitem[LeCun \& Huang, 2005][LeCun and Huang][2005]{lecun05loss}
LeCun, Y., \& Huang, F.J. (2005).
\newblock Loss functions and discriminitive training of energy-based
  models.
\newblock {\em AI-Stats}.

\vspace{-2mm}\bibitem[McAllester et~al.\/, 2004][McAllester et~al.\/][2004]{mcallester04cfd}
McAllester, D., Collins, M., \& Pereira, F. (2004).
\newblock Case-factor digrams for structured probabilistic modeling.
\newblock {\em UAI}.

\vspace{-2mm}\bibitem[McCallum et~al.\/, 2000][McCallum et~al.\/][2000]{mccallum00memm}
McCallum, A., Freitag, D., \& Pereira, F. (2000).
\newblock Maximum entropy {Markov} models for information extraction and
  segmentation.
\newblock {\em ICML}.

\vspace{-2mm}\bibitem[McCallum \& Wellner, 2004][McCallum and
  Wellner][2004]{mccallum04coref}
McCallum, A., \& Wellner, B. (2004).
\newblock Conditional models of identity uncertainty with application to noun
  coreference.
\newblock {\em NIPS}.

\vspace{-2mm}\bibitem[McDonald et~al.\/, 2004][McDonald et~al.\/][2004]{mcdonald04margin}
McDonald, R., Crammer, K., \& Pereira, F. (2004).
\newblock Large margin online learning algorithms for scalable structured
  classification.
\newblock {\em NIPS Workshop on Structured Outputs}.

\vspace{-2mm}\bibitem[Punyakanok \& Roth, 2001][Punyakanok and
  Roth][2001]{punyakanok01inference}
Punyakanok, V., \& Roth, D. (2001).
\newblock The use of classifiers in sequential inference.
\newblock {\em NIPS}.

\vspace{-2mm}\bibitem[Rosenblatt, 1958][Rosenblatt][1958]{rosenblatt58perceptron}
Rosenblatt, F. (1958).
\newblock The perceptron: A probabilistic model for information storage and
  organization in the brain.
\newblock {\em Psychological Review}, {\em 65}.

\vspace{-2mm}\bibitem[Russell \& Norvig, 1995][Russell and Norvig][1995]{russell95aibook}
Russell, S., \& Norvig, P. (1995).
\newblock {\em Artificial intelligence: A modern approach}.
\newblock New Jersey: Prentice Hall.

\vspace{-2mm}\bibitem[Sarawagi \& Cohen, 2004][Sarawagi and Cohen][2004]{sarawagi04scrf}
Sarawagi, S., \& Cohen, W. (2004).
\newblock Semi-{Markov} conditional random fields for information extraction.
\newblock {\em NIPS}.

\vspace{-2mm}\bibitem[Sha \& Pereira, 2002][Sha and Pereira][2002]{sha02crf}
Sha, F., \& Pereira, F. (2002).
\newblock Shallow parsing with conditional random fields.
\newblock {\em NAACL/HLT}.

\vspace{-2mm}\bibitem[Sutton et~al.\/, 2004][Sutton et~al.\/][2004]{sutton04fcrfs}
Sutton, C., Rohanimanesh, K., \& McCallum, A. (2004).
\newblock Dynamic conditional random fields: Factorized probabilistic models
  for labeling and segmenting sequence data.
\newblock {\em ICML}.

\vspace{-2mm}\bibitem[Sutton, 1996][Sutton][1996]{sutton96generalization}
Sutton, R. (1996).
\newblock Generalization in reinforcement learning: Successful examples using
  sparse coarse coding.
\newblock {\em NIPS}.

\vspace{-2mm}\bibitem[Taskar et~al.\/, 2003][Taskar et~al.\/][2003]{taskar03mmmn}
Taskar, B., Guestrin, C., \& Koller, D. (2003).
\newblock Max-margin Markov networks.
\newblock {\em NIPS}.

\vspace{-2mm}\bibitem[Tsochantaridis et~al.\/, 2004][Tsochantaridis
  et~al.\/][2004]{tsochantaridis04svmso}
Tsochantaridis, I., Hofmann, T., Joachims, T., \& Altun, Y. (2004).
\newblock Support vector machine learning for interdependent and structured
  output spaces.
\newblock {\em ICML}.

\vspace{-2mm}\bibitem[Weston et~al.\/, 2002][Weston et~al.\/][2002]{weston02kde}
Weston, J., Chapelle, O., Elisseeff, A., Schoelkopf, B., \& Vapnik, V. (2002).
\newblock Kernel dependency estimation.
\newblock {\em NIPS}.

\vspace{-2mm}\bibitem[Zhang et~al.\/, 2002][Zhang et~al.\/][2002]{zhang02chunking}
Zhang, T., Damerau, F., \& Johnson, D. (2002).
\newblock Text chunking based on a generalization of winnow.
\newblock {\em JMLR} {\em 2}.

\vspace{-2mm}\bibitem[Zhang \& Dietterich, 1997][Zhang and
  Dietterich][1997]{zhang97combinatorial}
Zhang, W., \& Dietterich, T.~G. (1997).
\newblock {\em Solving combinatorial optimization tasks by reinforcement
  learning: A general methodology applied to resouce-constrained scheduling}
  (Technical Report).
\newblock Oregon State University.

\vspace{-2mm}\bibitem[Zinkevich et~al.\/, 2005][Zinkevich et~al.\/][2005]{zinkevich05marginal}
Zinkevich, M., Riley, P., Bowling, M., \& Blum, A. (2005).
\newblock {\em Marginal Best Response, {Nash} Equilibria, and Iterated Gradient Ascent}.
\newblock In preparation.

\end{thebibliography}

\appendix
\vspace{-5mm}
\subsection*{Appendix A. Proof of Theorem 4}
\vspace{-1mm}

We follow \emcite{gentile-alma}, Thm 3, modifying the bound of the
normalization factor when projecting $\vec w$; suppose $\vec w$ is the
optimal separating hyperplane.  Denoting the normalization factor
$N_k$ on update $k$, we find: $N_{k+1}^2 \leq \norm{\vec w_k + \eta_k
\vec \De}^2 \leq \norm{\vec w_k}^2 + \eta_k^2 + 2 \eta_k \vec w_k \T
\vec \De \leq 1 + \eta_k^2 + 2(1-\al)\eta_k \ga$ ($\ga$ is the margin)
by observing $\vec \De$ is bounded above by $\ga$ since $\vec w\T
\left[\sum_{n \in \textit{sibs}} \vec \Ph(x,n) / |\textit{sibs}| -
\sum_{n \in \textit{nodes}} \vec \Ph(x, n) / |\textit{nodes}|\right]
\leq \vec w\T \left[\max_{n \in \textit{sibs}} \vec \Ph(x,n) - \min_{n
\in \textit{nodes}} \vec \Ph(x, n)\right] \leq \ga$, due to the
definition of the margin.  $N_k$ is bounded by $1+(8/\al-6)/k$ to
bound number of updates $m$ by $\ga m \leq
(4/\al-2)\sqrt{4/\al-3+m/2}$.  Algebra completes the proof.


\end{footnotesize}

\end{document}